\documentclass[%
 aip,
 amsmath,amssymb,
 reprint,%
]{revtex4-1}

\usepackage{graphicx}% Include figure files
\usepackage{dcolumn}% Align table columns on decimal point
\usepackage{bm}% bold math
\usepackage{diagbox}

\usepackage[utf8]{inputenc}
\usepackage[T1]{fontenc}
\usepackage{mathptmx}

\begin{document}

\preprint{AIP/123-QED}

\title[Deep Learning of Chaos Classification]{Deep Learning of Chaos Classification}
% Force line breaks with \\

\author{Woo Seok Lee}
\affiliation{Center for Theoretical Physics of Complex Systems, Institute for Basic Science, 34051 Daejeon, Korea}
\author{Sergej Flach}
\affiliation{Center for Theoretical Physics of Complex Systems, Institute for Basic Science, 34051 Daejeon, Korea}
\affiliation{Basic Science Program, Korea University of Science and Technology (UST), 34113 Daejeon, Korea}

\date{\today}% It is always \today, today,

\begin{abstract}
We train an artificial neural network which distinguishes chaotic and regular dynamics of the two-dimensional Chirikov standard map. 
We use finite length trajectories and compare the performance with traditional
numerical methods which need to evaluate the Lyapunov exponent. 
The neural network has superior performance for short periods with length down to 10 Lyapunov times on which the traditional Lyapunov exponent computation is far from converging. 
We show the robustness of the neural network to varying control parameters, in particular we train with one set of control parameters, and successfully  test in
a complementary set. Furthermore, we use the neural network to successfully test the dynamics of discrete maps in different dimensions, e.g. the one-dimensional logistic map and a three-dimensional discrete version of the Lorenz system. Our results demonstrate that a convolutional neural network can be used as an excellent chaos indicator.
\end{abstract}

\maketitle

\begin{quotation}
In low-dimensional dynamical systems with a mixed phase space, both regular and chaotic domains coexist. 
Lyapunov exponents characterise the time-averaged exponential divergence of nearby orbits in phase space, and are traditional chaos indicators
telling regular (zero exponent) from chaotic (nonzero exponent) dynamics apart.
In computational approaches, Lyapunov exponents are calculated through long time iterations as finite time averages of properly defined observables. 
To tell whether an orbit is regular and chaotic means to tell whether the finite time average tends to zero or nonzero values in the infinite time limit.
Such traditional methods therefore suffer from notorious uncertainties and need iteration times which are orders of magnitude longer than
typical Lyapunov times (the inverse of the exponents).
%to accomplished the exponential values to converge. 
On the other side deep learning algorithms turned to be useful to train networks which then serve as efficient classifiers. 
These can be used to study heartbeat irregularities, weather forecasting, complex dynamics reconstruction, pattern recognition and feature extraction,
among many others.
When using in particular a convolutional neural network, the present work shows that
% has proven to be useful in chaos system analysis, such as heartbeat irregularities and weather forecasting~\cite{skinner1990, slingo2011}. In addition, deep learning algorithms are used in various fields, which are known as powerful tools for pattern recognition and feature extraction. In this paper, we build a deep learning model for chaos indicator using the simple deep convolutional neural network~\cite{LeCun}. We show that this model 
a novel classifier emerges, which operates at high accuracy yet uses finite time orbits which are orders of magnitude shorter than the
ones needed for the same accuracy with traditional Lyapunov methods. 
%
%is able to distinguish chaos trajectories with high accuracy using short limited time series measurements. It also suggests the possibility that this model can be used as a general chaos indicator by showing that a model trained with a particular chaos system has some degree of chaos discrimination ability in other chaos systems.
\end{quotation}

\section{Introduction}

Chaotic dynamics exists in many natural systems, such as heartbeat irregularities, weather and climate \cite{skinner1990, slingo2011}. Such dynamics can be studied through the analysis of proper mathematical models which generate nonlinear dynamics and determenistic chaos. Chaotic and regular dynamics can co-exist in the phase space of low-dimensional systems  \cite{ott2002chaos}.
%These models have chaos or regular trajectory according to specific parameter values. 
To distinguish chaotic from regular dynamics, the tangent dynamics is used to compute Lyapunov exponents $\lambda$. In practice one integrates the tangent dynamics along a given
trajectory and averages a finite time Lyapunov exponent $\lambda(t)$. The averaging time $T$ needed to reliably tell regular ($\lambda=0$) from 
chaotic ($\lambda \neq0$) trajectories apart is usually orders of magnitude larger than the Lyapunov time $T_{\lambda} \equiv 1/\lambda$.
%In mathematics, the Lyapunov exponent of the dynamical system is the separation rate of an infinitely close trajectory. If all points in a neighborhood of trajectory converge toward the same orbit, it is a fixed point or limit cycle. However, if the system has chaotic behavior, the distance between any two trajectories $x_n=f^{n}(x_0)$ and $x_n+\delta x_n=f^{n}(x_0+\delta x_0)$ separate exponentially with time. The problem with calculating a Lyapunov exponent is that we need to know evolution of two close trajectories. For experimental data that do not know the exact dynamics, this is a major obstacle. Another problem is that it takes a long time iteration to calculate the correct lyapunov exponent. 

%{\bf \color{red} REVISED UP TO HERE}

Here, we introduce a machine learning approach that alleviates the problems of calculating Lyapunov exponents and can be used as a new chaos indicator.
% Chaos systems are very sensitive to their initial conditions. Small initial condition differences in a chaos system ultimately lead to a completely different trajectory. One way to distinguish chaos trajectory is to calculate Lyapunov exponent. It merely quantifies the degree of "sensitivity to initial conditions". However, we need a lots of time step to calculate exact Lyapunov exponent.
% Another property of chaotic system is that it is deterministic, meaning that their trajectories are determined by the system unlike a stochastic system. When the initial value is determined in the chaos system, the trajectory is determined, but the system shows a completely different trajectory when the initial value is changed slightly~\cite{kellert1993}.
Machine learning has shown tremendous performance e.g. in pattern recognition \cite{dodge2017study, al2017review}. Machine learning approaches turned useful to solve partial differential equations and identify hidden physics models from experimental data~\cite{rudy2017,han2018, raissi2018}. 
Machine learning was used recently to predict future chaotic dynamics details from time series data without knowledge of the generating equations~\cite{agrawal2019, pathak2018}. In this paper, we introduce a machine learning way to use short time series data for telling chaos from regularity apart.
We train a neural network using chaotic and regular trajecories from the Chirikov standard map. Our method has a success rate of 98\% using
trajectories with length $10T_{\lambda}$, while conventional methods need up to $10^4 T_{\lambda}$ to reach the same accuracy. The main reason for the small but
finite failure rate of our machine learning method is due to sticky orbits. These orbits are chaotic,yet can mimic regular ones for long times due to trapping
in fractal boundary phase space regions separating chaotic and regular dynamics. Our method is also surprisingly successfull when trained with Standard Map data but
tested on maps with different dimensions such as the logistic map ($d=1$) and the Lorenz system ($d=3$).

\section{The Chirikov Standard Map}
The Chirikov standard map is an area-preserving map in dimension $d=2$ \cite{lichtenberg2013regular} also known as the kicked rotor \cite{ott2002chaos}
%The time evolution of the system is described by the dimensionless Hamiltonian
%{\bf \color{red} references to chirikov and LL from wikipedia}
:
%\begin{equation}
%\begin{aligned} 
%\label{standard_map_hamiltonian}
%    \mathcal{H}(p,x,t)=\frac{p^2}{2}+\delta_{1}(t)Kcos(x)
%\end{aligned}
%\end{equation}
%where $\delta_1(t)$ is the 1-periodic Dirac-delta, which can be reformulated as a two dimensional discrete map:
\begin{equation}
\begin{aligned} 
\label{standard_map}
    p_{n+1}=p_n+\frac{K}{2\pi}sin(2\pi x_n) \qquad mod\ 1 \;,\\
    x_{n+1}=x_n+p_{n+1} \qquad  mod\ 1 \;. \\
\end{aligned}
\end{equation}
The kick strength $K$ controls the degree of nonintegrability and chaos appearing in the dynamics generated by the map.
\begin{figure} [hbt!]
	\centering
	\includegraphics[width=\columnwidth]{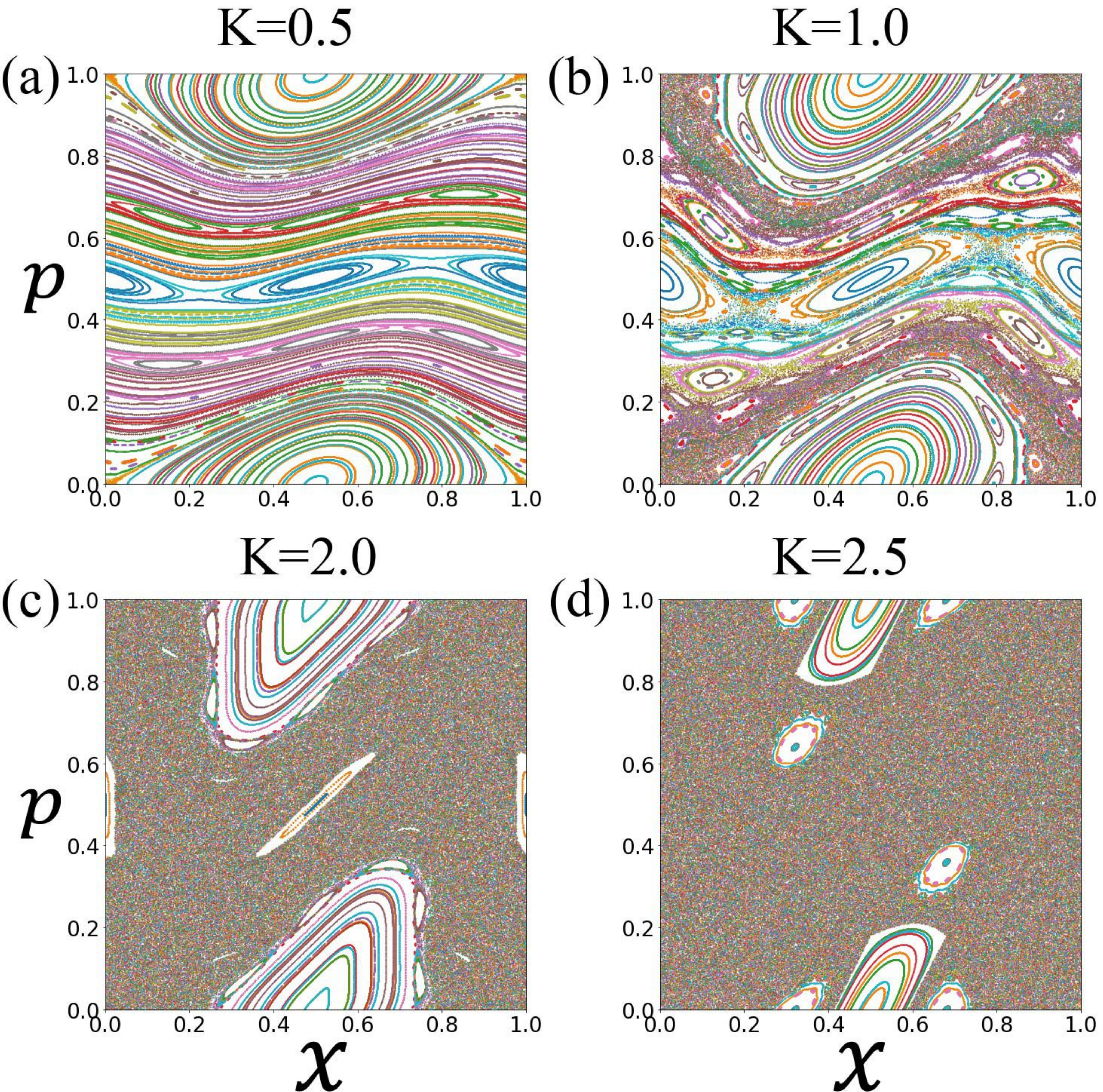}
	\caption{\label{standardmap_poincare}Examples of Poincare sections of the standard map. (a) K=0.5, (b) K=1.0, (c) K=2.0, (d) K=2.5.
% {\bf \color{red} Remove the margins in the figure.}
}
\end{figure}
%This model is known as the Kicked Rotator~\cite{ott2002chaos}. $p$ and $x$ can be considered to respectively represent the position and momentum of a pendulum with as impulsive torque $K sin(x)$ at the end of the pendulum. $K$ represents the amplitude of the impulsive torque and determines the extent of nonlinearity of the system. 
Consider the case when $K=0$. Eq.~\ref{standard_map} reduces to $p_{n+1}=p_n \quad (mod\ 1)$ and $x_{n+1}=x_n+p_{n+1} \quad  (mod\ 1)$ which is integrable and every orbit resides on an invariant torus. The orbit can exhibit periodic or quasi-periodic behavior depending on the initial conditions ($p_0, x_0$). 
For small values of $K$ e.g. $K=0.5$ (Fig.\ref{standardmap_poincare}(a)) most of these orbits persist, with tiny regions of chaotic dynamics appearing which 
are not visible on the presented plotting scales.
At $K=K_c\approx 0.97$ the last invariant KAM tori are destroyed and a simply connected chaotic sea is formed which allows
for unbounded momentum diffusion.
For larger values of $K$ the chaotic fraction grows confining regular dynamics to regular islands embedded in a chaotic sea (Fig.~\ref{standardmap_poincare}).
Further increase of $K$ leads to a flooding of the regular islands by the chaotic sea.

%As can be seen in Fig.~\ref{standardmap_poincare}. the non-chaotic (regular) trajectory exists only within a certain area. I will refer to such area as "island" that are never visited during iteration. For small values of $K$ there are trajectories that split the phase space into disjoint manifolds. As $K$ increas, the nonlinearity of the system increases. We can see the region where the invariant trajectories alive, is slowly swallowed up with increasing $K$ (Fig.~\ref{standardmap_poincare}).

% {\color{red} introduce Fig.1. Add also a Poincare map with only black points for both cases $K=1,2$ which shows
% the phase space structure in more detail}

\section{Lyapunov exponents and predictions}
The Lyapunov exponent (LE) characterizes the exponential rate of separation of a trajectory $\{p_n,x_n\}$ and its infinitesimal perturbation $\{\delta_n,\zeta_n\}$:
%two infinitesimally separated trajectories. The trajectory starting at a position slightly away from the initial position $(p_{0},x_{0})$ is as follow.
% The Lyapunov exponent is typically computed by following the linearization along a given reference trajectory.
\begin{equation}
\begin{aligned} 
\label{nearby_map}
    p_{n+1}+\delta _{n+1}&=(p_n+\delta _{n})+\frac{k}{2\pi}sin(2\pi (x_n+\zeta _{n})) \\
    x_{n+1}+\zeta _{n+1}&=(x_n+\zeta _{n})+(p_{n+1}+\delta _{n+1}) \\
\end{aligned}
\end{equation}
Linearizing (\ref{nearby_map}) in the perturbation yields the tangent dynamics generated by the variational equations
%The linear approximation of Eq.~\ref{nearby_map} and Eq.~\ref{standard_map} yield equations for the distance between two trajectories, as shown in Eq.~\ref{tan_map}
\begin{equation}
\begin{aligned} \label{tan_map}
    \delta _{n+1}&=\delta _n + k \zeta _{n} cos(2\pi x_{n}) \\
    \zeta _{n+1}&=\zeta _n + \delta _{n+1} \\
\end{aligned}
\end{equation}
For computational pruposes
$\delta$ and $\zeta$ can be rescaled after any time step without loss of generality, while keeping the rescaling factor. The LE $\lambda$ for each trajectory is obtained from the
time dependence of $\lambda_N$: 
\begin{equation}
%\begin{aligned} 
\label{lyapunov_exp}
    \lambda_{N}=\frac{1}{N}\sum_{n=2}^{N}\ln (\frac{\sqrt{\delta_{n}^{2}+\zeta_{n}^{2}}}{\sqrt{\delta_{n-1}^{2}+\zeta_{n-1}^{2}}}) \; , \;
\lambda = \lim_{N\rightarrow \infty} \lambda_N \;.
% ~ \lambda_{N}=1/Z_{N}
%\end{aligned}
\end{equation}
% \begin{figure} [ht]
% 	\centering
% 	\includegraphics[width=\columnwidth]{FWHM.pdf}
% 	\caption{\label{FWHM}  (a) Overlap coefficient and (b) FWHM in Multi-Gaussian fitting of Lyapunov exponent histogram in Fig.~\ref{histogram}.}
% \end{figure}
%$\lambda_{N \rightarrow \infty} \equiv \lambda$ for the main chaotic layer.  
The Lyapunov time is then defined as $T_{\lambda} \equiv 1/\lambda$. For the main chaotic sea it is a function of the control parameter $K$. A suitable fitting function yields 
%and takes values {\bf \color{red} we need either a figure, or a table, and/or a formula!!!}
%
%{\bf \color{blue}In the standard map, the Lyapunov exponent of the chaotic sea varies with the nonlinearity parameter K in Eq.~\ref{tan_map}. In other words, the numerical value of $\lambda_{N}$ can be expressed as a function of K. The function 
$\lambda \approx \ln(0.7+0.42K)$ 
%is a suitable fitting function for $20>K>K_{c}$.~
\cite{harsoula2019characteristic}.
% Here, $K_{c}$ is a critical value and when $K < K_{c}$, chaotic areas do not communicate with each other in phase space.
%}

\begin{figure} [ht]
	\centering
	\includegraphics[width= 0.99 \columnwidth]{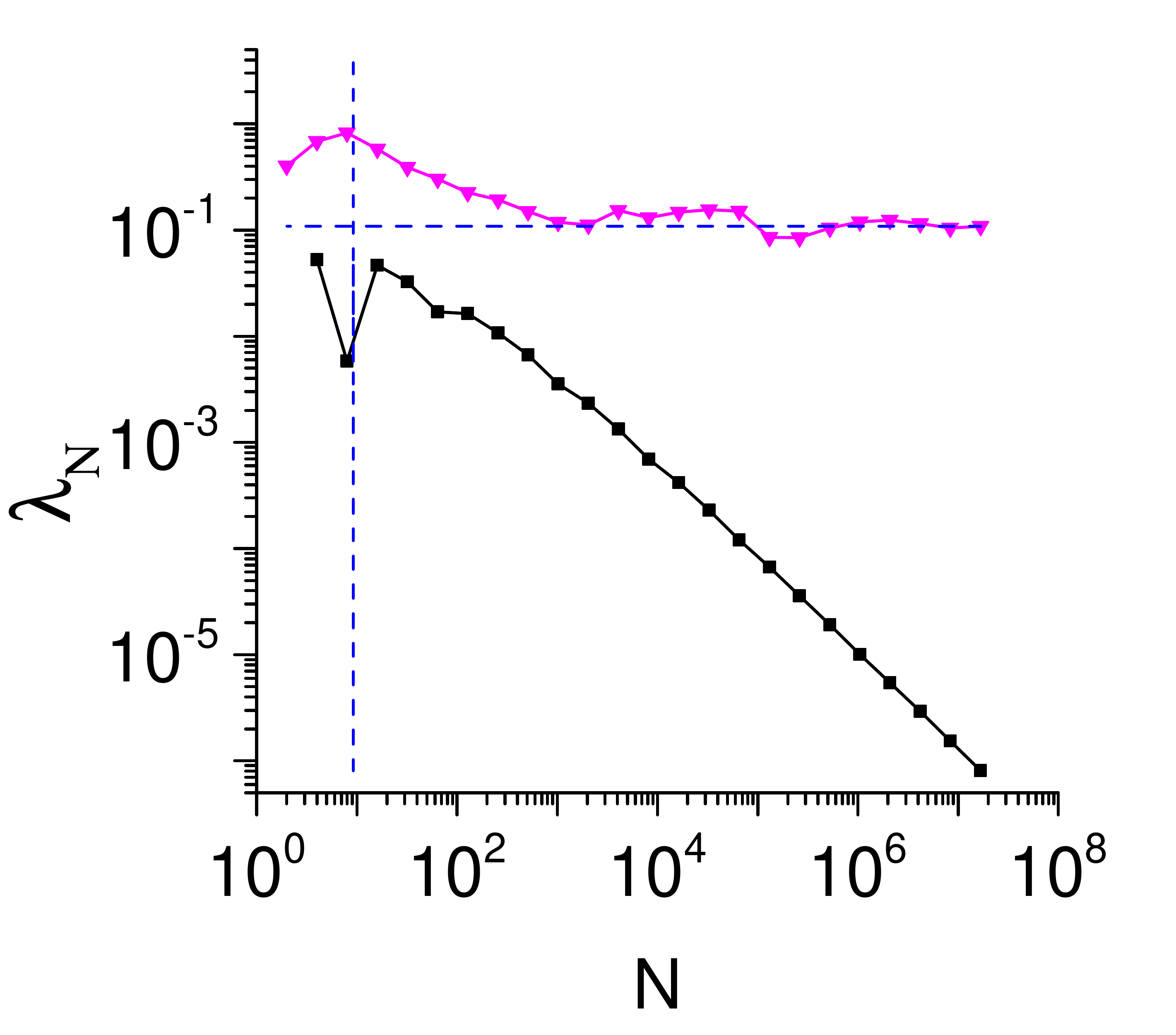}
	\caption{\label{K1_0} 
$\lambda_N$ versus $N$ for a chaotic (triangles) respectively regular (squares) trajectory with $K=1.0$. 
The dashed horizontal line indicates the value of $\lambda$ for the chaotic trajectory, and the dashed vertical one the 
corresponding value of $T_{\lambda}$.
}
\end{figure}

For a regular trajectory $\lambda_N \sim 1/N$ and $\lambda=0$, at variance to a chaotic trajectory for which $\lambda_N$ saturates at $\lambda$
at a time $N\approx T_{\lambda}$. Technically this saturation, and the value of $\lambda$ can be safely confirmed and read off
only
on time scales $N \approx 10^2..10^3 T_{\lambda}$, without becoming a quantifiable distinguisher of the two types of 
trajectories, see Fig.\ref{K1_0}.

\begin{figure} [ht]
	\centering
	\includegraphics[width= 1.05 \columnwidth]{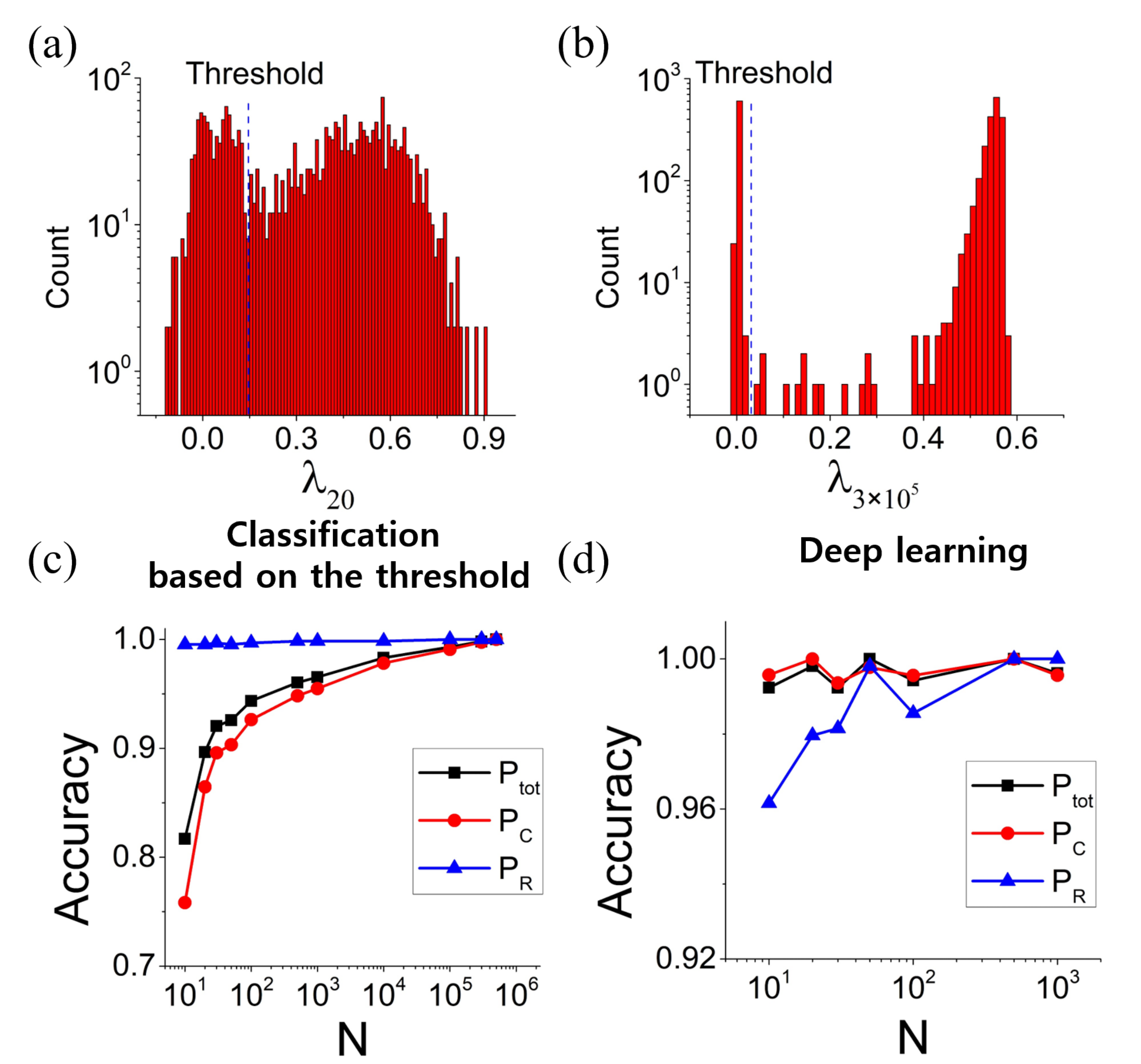}
	\caption{\label{histogram} Performance comparison of a Lyapunov exponent based method and a deep learning method to distinguish chaotic and regular trajectories for $K=2.5$ and $\lambda \approx 0.56$.
(a) 
Histogram of of $\lambda_{N=20}$. the dashed vertical line indicates the location of the threshold (see text for details).
(b) Same as (a) but $N=3\times 10^5$.
(c) The success rates $P_R$, $P_C$ and  $P_{tot}$ as a function of $N$ for the Lyapunov exponent based method (see text for details).
(d) Same as in (c) but for the deep learning based method.The network was trained for  K = 2.5 and 2081 trajectories. The remaining 520 trajectories are used for testing.  $N$ in (d) represents the trajectory length used for network training and test. $K_{min}=K_{max}=2.5$, $M_{tr}=2081$, $M_{tt}=520$, $N_{K}\equiv N$}
\end{figure}

To quantify our statements, we run the standard map at $K=2.5$ Fig.\ref{standardmap_poincare}(d). We use a grid of $51 \times 51$ points which partitions
the phase space $\{p,x\}$ into a square lattice. We use the corresponding 2601 initial conditions and generate trajectories. Each trajectory returns a function
$\lambda_N$. We plot the resulting histogram for $N=20$ and $N=3\cdot10^5$ in Fig.\ref{histogram} (a) and (b) respectively. 
For $N\rightarrow \infty$ the histogram should show two bars only - one at $\lambda_N=0$ (all regular trajectories) and one at $\lambda_N=\lambda$ 
(all chaotic trajectories). For finite $N$ the distributions smoothen. Note that even negative values $\lambda_N$ are generated due to fluctuations and finite
averaging times. To tell chaotic from regular dynamics apart, we use the following protocol. We identify the two largest peaks in each histogram, and identify the
threshold dividing dynamics into regular and chaotic as the deepest minimum between them (in case of a degeneracy, the one with the smallest value of $\lambda_N$). 
The location of the threshold is shown for $N=20$ and $N=3\cdot10^5$ in Fig.\ref{histogram} (a) and (b) respectively. 
We then assign a chaos respectively regular label to each trajectory. This label can fluctuate as a function of time for any given trajectory.
We use the division for the largest simulation time $N=3 \cdot 10^5$ as a reference ('true') label for all trajectories. 
The success rate in predicting
the correct regular $P_R$ or chaotic $P_C$ label is defined by the ratio of the correctly predicted labels within each subgroup of identical true labels. 
Likewise the success rate of predicting any label correctly is denoted by $P_{tot}$.
The results are plotted versus time $N$ in Fig.\ref{histogram} (c). While regular labels are predicted with high accuracy,
chaotic ones are reaching 98$\%$ at only $N \approx 10^3 T_{\lambda}$. 
%{\bf \color{red} Check once we have the proper numbers for $T_{\lambda}$!}. 
The low success rate $P_C$ is therefore also lowering the total success rate $P_{tot}$.

\section{Neural networks and predictions}

%{\bf \color{red} We need to introduce the network in general terms for the non-specialist here. All technical details should be outsourced into the appendix,
%including the detailed figure (if needed). A more schematic general figure could be instead presented in this main text.}

%\subsection{Convolution neural network}

The input data of an artificial neural network consisting of only fully connected layers are limited to a one-dimensional (array) form~\cite{ramsundar2018tensorflow}.
Fully connected layers connect all the inputs from one layer to every activation unit of the next layer.
The standard map generates sequences embedded in two dimensions.
In order to learn data embedded in dimensions two or larger, the data must be flattened, and spatial information can get lost. 
A Convolutional Neural Network (CNN) is known to learn while maintaining spatial informations of images \cite{LeCun}.
A CNN is usually configured with convolution and pooling layers. The former employ convolutional integrals with input data and filters to produce output feature maps. An additional activation function turns the network non-linear. At the end of the convolution layers a pooling layer is added which performs value extraction in a given pooling region. Through multiple convolution layers and pooling layers, the network can improve
its prediction features. 
%Because of the {\bf \color{red} you did not define 'global max pooling layer', and the reader will not know it either. Above we only loosely defined 'pooling layers'. Modify!} global max pooling layer, the extracted features are arranged in a one-dimensional array.  
Finally, a fully connected layer generates classified output data. 
For binary classification, the last layer consists of one node. Its output value is either zero or one.
We refer the reader to Appendix \ref{app1}
for further technical details of the CNN we use.

\subsection{The standard map}
The input of the neural network is a time series $(p_{n},x_{n})$ from Eq.~\ref{standard_map}. The trajectory $(p_{n},x_{n})$ shows regular or chaotic behavior depending on the initial values $(p_{0},x_{0})$. Each of the trajectories is assigned a class label based on the Lyapunov time: Class $R$ corresponds to a non-chaotic trajectories while $C$ corresponds to a chaotic trajectories.
We remind that the phase space is discretized into $51\times 51 =2601$ grid points. 
The training and testing is quantified with a set of parameters: 
i) $K_{min}$ and $K_{max}$ denote the range of training values of $K$ on an equidistant grid with $M_K$ values;
ii) $M_{tr}$ is the number of training trajectories per $K$ value;
iii) $N_K$ is the training trajectory length;
iv) $M_{tt}$ is the number of test trajectories per $K$ value.

To quantify the CNN performance, we assign a discrete label to each of the initial phase space points  - $C$ respectively $R$ based on
the Lyapunov exponent method with trajectory length $N=3 \cdot 10^5$. 
This way we separate all phase space points into two sets - $C$ and $R$, each containing $A_C$ and $A_R$ points. We then run the CNN prediction on trajectories
of length $N=20$ which start from each of the gridded phase space points. We compute the accuracy quantifying probabilities
%
%The end of the network is a softmax layer, which gives the output vector as a probability ($O_{R}, O_{C}$), where $O_{R}$ is the probability that the input vector is regular trajectory and $O_{C}=1-O_{R}$ is the probability that it is chaos. A input is classified as regular when $O_{R}>0.5$ and chaos otherwise. Since the chaos region changes according to the K value at Eq.~\ref{standard_map}, 
%the performance of the neural network is determined using chaos and regular region accuracy defined as 
\begin{equation}\label{accuracy}
P_{C} = \frac{B_{C}}{A_{C}}, ~
P_{R} = \frac{B_{R}}{A_{R}}, ~
P_{tot}= \frac{B_{C}+B_{R}}{A_{C}+A_{R}}
\end{equation}
%where $A_{C}$ and $A_{R}$ are the numbers of chaos and regular trajectories respectively, which 
where $B_{C}$ and $B_R$ are the numbers of trajectories predicted by the CNN to be chaotic respectively regular within each
of the true sets $A_C$ and $A_R$. Thus strictly $B_C \leq A_C$ and $B_R \leq A_R$.

%is the number of trajectory that the model predicted as chaos in $A_{C}$ ($B_{C}\leq A_{C}$). Similarly $B_{R}$ is the number of trajectory that the model predicted as regular in $A_{R}$ ($B_{R}\leq A_{R}$).

\begin{figure} [hbt!]
	\centering
	\includegraphics[width=\columnwidth]{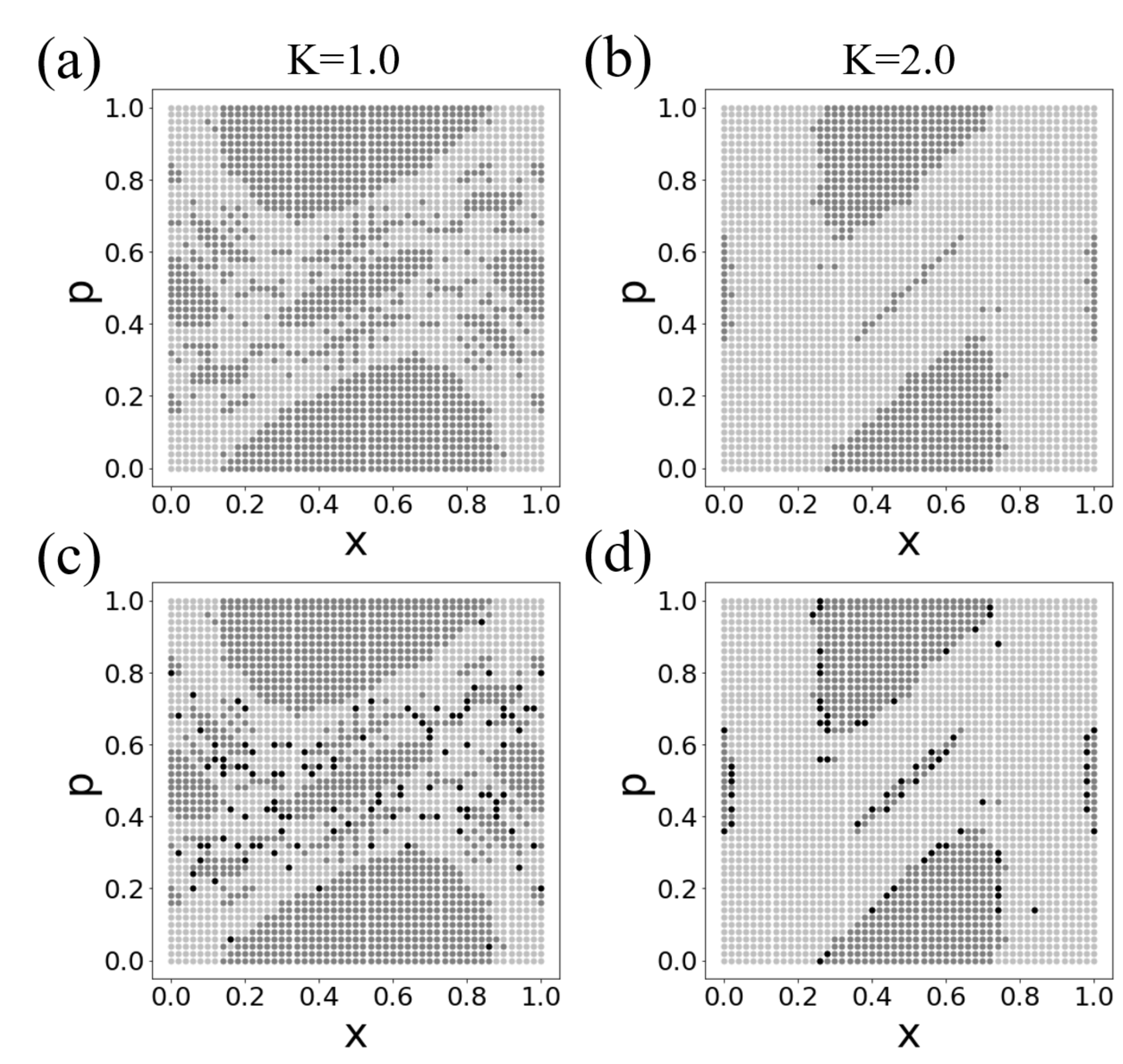}
	\caption{\label{standardmap_lyapunov} 
%{\bf \color{red} WHAT IS THAT?}
Chaos classification in the standard map. 
The Lyapunov exponent classification with trajectory length $N=3 \cdot 10^5$
is used as a reference classifyer for $K=1$ (a) and $K=2$ (b). 
The CNN test results are shown for $K=1$ (c) and $K=2$ (d).
Open circles - regular, gray circles - chaotic. 
Black circles show the error locations of the CNN prediction.
The CNN parameters are $K_{min}=1.0$, $K_{max}=2.0$, $M_K=11$, $M_{tr}=2081$, $M_{tt}=520$, $N_K=10$.
% (a),(b) and a convolutional neural network with $N=20$ (c),(d).
%{\color{blue}{Open circles - regular, gray circles - chaotic. $K=1.0$ in (a),(c) and $K=2.0$ in (b),(d). black circles indicate errors of the neural network prediction. $K_{min}=1.0$, $K_{max}=2.0$, $M_K=11$, $M_{tr}=2081$, $M_{tt}=520$, $N_K=20$.}}
%{\bf \color{red} Add all necessary parameters which quantify the CNN results $K_{min},K_{max},M_K,M_{tr},M_{tt},N_K$.}
%
%
%Chaotic and regular regions as obtained from 
%$\lambda_{N}$ in the standard map at $K=1.0$ and (b) $K=2.0$ ($N=300,000$). The red dots are regular and the blue dots are chaotic regions. (c) ,(d) Prediction results using the model in Table1 ($10T_{\lambda_{3\times10^5}}$). Black dots indicate errors.
}
\end{figure}

Fig. \ref{histogram}(d) compares the CNN performance to the standard Lyapunov base one.
Accuracies of 98\% and more are reached by the CNN for trajectory length $N_K \geq 30$. Similar accuracies
need trajectory length $N \approx 10^4$ and more when using standard Lyapunov testing.
Fig.\ref{standardmap_lyapunov} shows the CNN performance with $N_K=10$ in the phase space of the standard map. 
We observe that most of the failures correspond to chaotic trajectories starting in the fractal border region close
to regular islands. These trajectories can be trapped for long times in the border region, with trapping time distributions
exhibiting power law tails \cite{zaslavsky1998physics}.
%and compares the chaos classification using a Lyapunov exponent based method (see previous section) for trajectories of length $N=3\cdot 10^5$ and a CNN based method with trajectory length $N=10 T_{\lambda}$. 
To quantify the performance of the CNN, we first vary the $N_K$ from 1 to 20  (Table~\ref{accuracy_table}).
The network is trained with chaotic and regular trajectories for $K_{min}=1.0$, $K_{max}=2.0$, $M_{K}=11$, and $1 \leq N_K \leq 20$  and the network performance is evaluated for $3 \leq K \leq 3.5$ and $M_{K}=6$. The CNN requires that the length of test trajectories is always kept equal to the length of the training trajectories.
%{\bf \color{blue}{( The length of test trajectory should be the same with training length because the size of the input layer is determined by the length of the training trajectory, and the weight matrix size between the input and the hidden layer varies depending on the size of the input layer even if I fix the hidden layer size. For example, if input layer size is 2 and hidden layer size 4, the weight matrix which represent weighted connection between input and hidden layer is $2 \times 4$ matrix but if I change the input size 2 to 3, the weight matrix size is $3 \times 4$ matrix.)}}
Note that the Lyapunov time $T_{\lambda} \approx 2$ for the test values of $K$.
The CNN shows improvement of the accuracy with increasing $N_K$.  While the performance fluctuates with varying $K$, it shows excellent results for $N_K$ values and clearly outperforms the Lyapunov exponent based method.
%{\color{blue}{The CNN distinguishes more than 90\% of chaos from regular trajectory using only $N_K=20$. Because of the sticky trajectory, the network classifies chaos trajectory as regular or regular trajectory as chaos. In addition, even with only $N_K=2$, the network shows near 90\% accuracy. Since the standard map has chaos and regular values determined by the initial values and K in Eq.~\ref{standard_map}, this short time trajectory can be considered to depend strongly only on the initial value. When training the network with $N_K=1$, the network converges to one map of the trained K values and always classifies a specific area as chaos and regular regardless of K values. As can be seen briefly in Fig.~\ref{standardmap_lyapunov}, as the K value increases, the size of the regular area decreases. Therefore, the network trained with $K_{min}=1.0$ and $K_{min}=2.0$ becomes $P_R=1$ for $K_{min}=3.0$ and $K_{min}=3.5$. This is obvious because the initial value does not contain information about K. It can be said that when $N_K=2$, the accuracy increased compared to when $N_K=1$, so network trained the information of K value included in the two iterations.}}
%The $T_{\lambda_{3\times10^5}}$ is approximately 2 when $K =3.0+0.1(m-1) \: (m\in \mathbb{Z}, 1 \leq m \leq 5)$. The length of trajectory used as network input varies from 2 to 20. (Table. ~\ref{accuracy_table}).
%{\bf \color{red}  Add all necessary parameters which quantify the CNN results $K_{min},K_{max},M_K,M_{tr},M_{tt},N_K$.   }
\begin{table*}[t]
\centering
% \begin{tabular}{|c|c|c|c|c|c|c|c|}
\begin{tabular}{c|c|c|c|c|c|c|c|c} 
\hline
\diagbox[width=11em]{K}{$N_K$} & $20$ & $18$ & $16$ & $14$ & $12$ & $10$ & $2$ & $1$  \\\hline
~ & $P_{C}$/$P_{R}$ & $P_{C}$/$P_{R}$ & $P_{C}$/$P_{R}$ & $P_{C}$/$P_{R}$ & $P_{C}$/$P_{R}$ & $P_{C}$/$P_{R}$ & $P_{C}$/$P_{R}$ & $P_{C}$/$P_{R}$\\
% \hline
3.0  &  0.99/0.99  & 0.93/0.98  & 0.95/0.98  & 0.92/0.98 & 0.97/0.96 &0.83/0.95&0.89/0.97&0.78/1.0 \\
3.1  &  0.90/0.98  & 0.94/0.96  & 0.96/0.96  & 0.93/0.96 & 0.90/0.96 &0.83/0.91&0.90/0.93&0.79/1.0 \\
3.2  &  0.93/0.95  & 0.94/0.97  & 0.96/0.97  & 0.93/0.97 & 0.97/0.94 &0.85/0.91&0.90/0.92&0.79/1.0 \\
3.3  &  0.97/0.99  & 0.93/0.99  & 0.95/0.99  & 0.93/0.99 & 0.94/0.96 &0.85/0.93&0.89/0.98&0.77/1.0 \\ 
3.4  &  0.94/0.99  & 0.89/0.97  & 0.94/0.96  & 0.92/0.98 & 0.93/0.97 &0.82/0.93&0.88/0.98&0.76/1.0 \\
3.5  &  0.93/0.94  & 0.93/0.93  & 0.96/0.88  & 0.92/0.99 & 0.92/0.91 &0.83/0.92&0.87/0.94&0.76/1.0 \\
\hline
\end{tabular}
\caption{CNN performance. For each K value, 2601 different initial values ($p_{0,i}, x_{0,j}$) were selected as ($p_{0,i}=(i-1)\frac{1}{50}, x_{0,j}=(j-1)\frac{1}{50},~ (i,j\in \mathbb{Z},\:1 \leq i,j \leq 51,\; )$). Other parameters are listed
in the main text.
}
\label{accuracy_table}
\end{table*}

\begin{figure} [ht]
	\centering
	\includegraphics[width=\columnwidth]{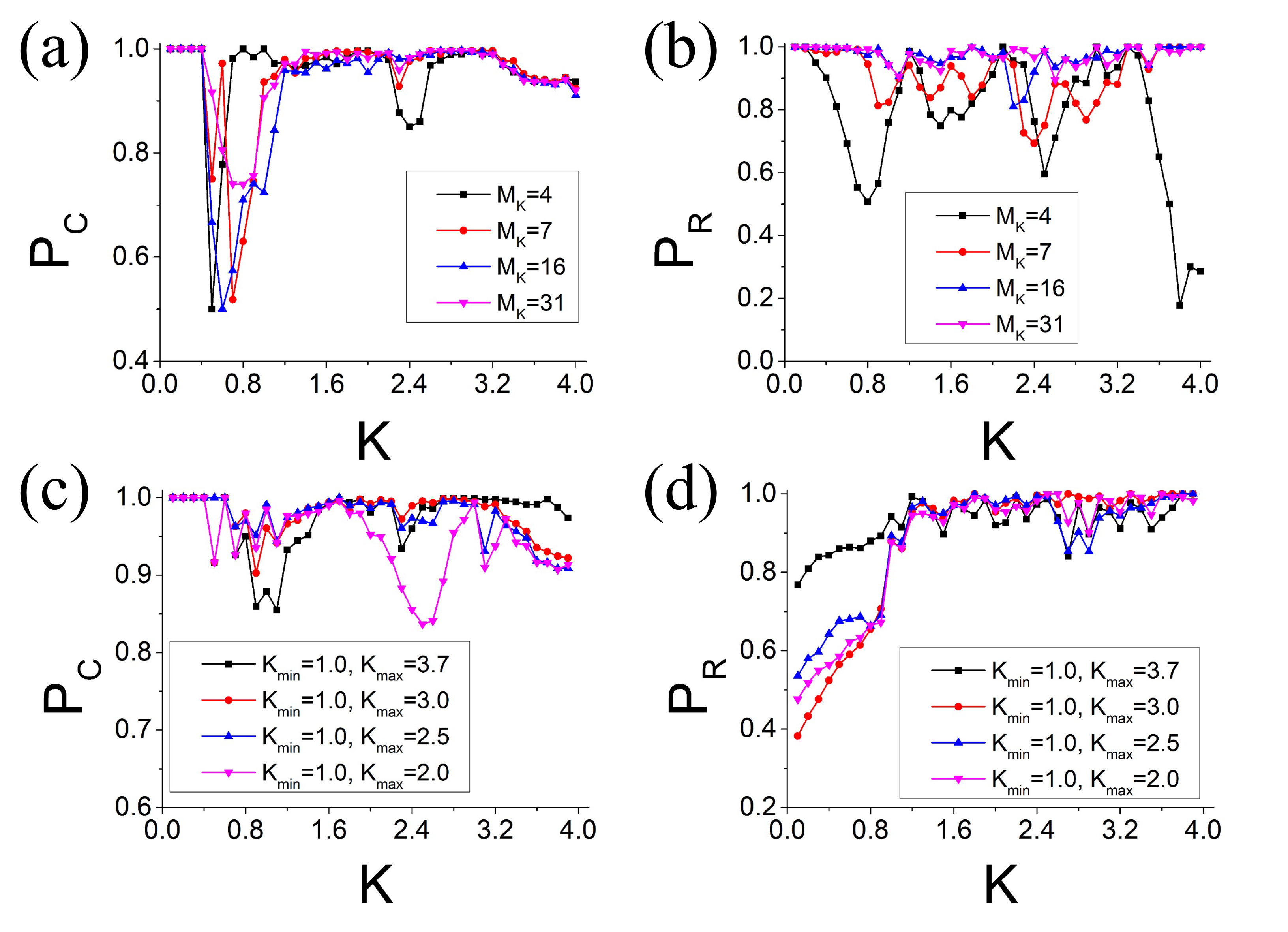}
	\caption{\label{Knum} 
Network performance versus $K$ for different trained K value numbers and ranges. (a), (b) Varying the number of K values used for network training in a fixed interval with equidistant spacing ($K_{min}=0.1,\;K_{max}=3.1\;M_{tr}=2081,\;M_{tt}=2601,\;N_K=20$). (black square) $M_K=4$. (red circle) $M_K=7$. (blue triangle) $M_K=16$. (magenta inverted triangle) $M_K=31$.
(c), (d)  Varying the interval of trained K values. The range of K values used in network learning  are (black square) $K_{min}=1.0,\;K_{max}=3.7,\;M_K=28$, (red circle) $K_{min}=1.0,\;K_{max}=3.0,\;M_K=21$, (blue triangle) $K_{min}=1.0,\;K_{max}=2.5,\;M_K=16$,(magenta inverted triangle) $K_{min}=1.0,\;K_{max}=2.0,\;M_K=11$. The length of the input trajectories are $20$.}
\end{figure}
We then further test the CNN performance for untrained $K$ values by varying the training $K$ range and other
relevant training parameters in Fig.~\ref{Knum}.
%with different trained $K$ value numbers and ranges.
%We examined how the range and number of K values used in network training affect network performance. 
The network shows better performance on untrained K values when trained with a set of different K values. 
%As shown in Fig.~\ref{Knum} (a), when comparing the performance of networks in the same K range, training the network with more K values shows better performance. 
As expected, smaller numbers of training $K$ values yield poorer accuracy due to overtraining.
%
%If only four K-values are used, the network accurately predict only the K-values trained, and the accuracy drops dramatically. The reason of this phenomenon is that the network is over-trained to a regular region of K = 1.0 and always predicts this area as a regular trajectory. As the K value increases, the regular region decreases and the chaos region increases, so if the network predicts that the regular area corresponding to K = 1.0 is regular regardless of the K value, the chaos accuracy decreases but the regular accuracy does not change. 
With increasing training range of $K$ values and ranges
%When the network trains a wide range of K values,
 the network improves its chaos region predictions for untrained K values. 
%In addition, as shown in Fig.~\ref{Knum} (b), when we increased the K value for network training by 0.1 at equal intervals, the larger the range, we have the higher the accuracy of the untrained range of K = 0.1 -- 0.9.
% {\color{red} Replace Fig.5 by Fig.7}.

%\begin{figure} [ht]
%	\centering
%	\includegraphics[width=\columnwidth]{Krange.pdf}
%	\caption{\label{Krange} Network accuracy according to training K value range. The (a), (b), and (c) are the result of training the network using K = 0.1, 0.2, ..., 1.0, K = 1.0, 1.1, ...,2.0, K = 2.0, 2.1, ..., 3.0 respectively. The length of the input is 10$T_{\lambda}$.}
%\end{figure}

%As a result of training the network by changing the range of K value, the huge error occurs at K values smaller than the trained K values (Fig~\ref{Krange}). When training with K = 0.1 ~ 1.0, which is a large error area, the network shows the large error even in the trained area (Fig~\ref{Krange}(a)).

%{\bf \color{red} WORKED ON TEXT UP TO HERE. DO NOT REMOVE MARKER}

\subsection{Training with the standard map, testing the logistic map}
\begin{figure} [ht]
	\includegraphics[width=200 pt]{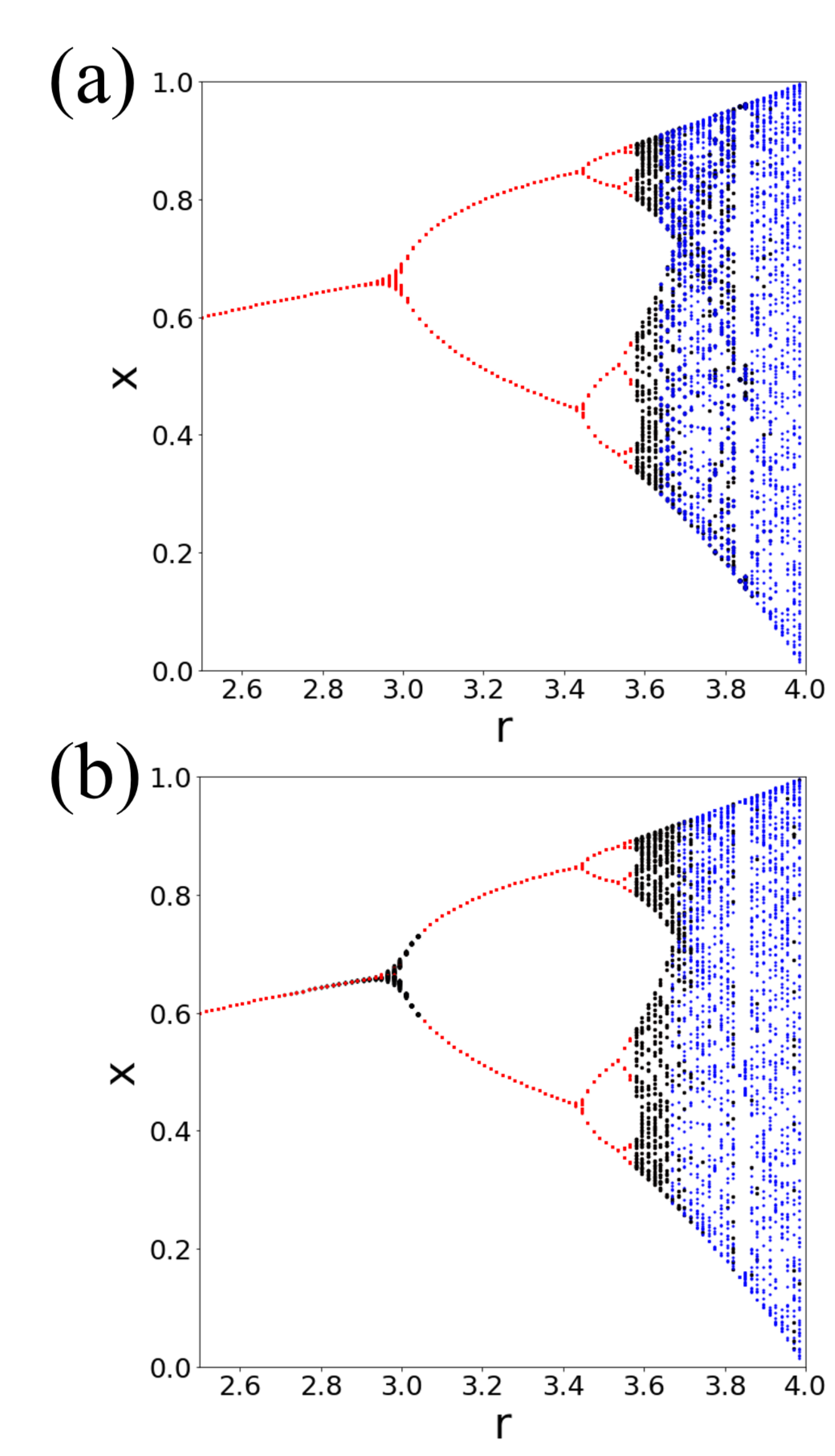}
	\caption{\label{logistic} The result of predictions for the logistic map with a network trained from the standard map. The blue and red dots are the cases where the network correctly predicts chaotic and regular trajectories respectively. The black dots show where the prediction fails. The network is trained with $K_{min}=1.0,\; K_{max}=2.0,\; M_K=11,\; M_{tr}=2081,\; M_{tt}=520,\;$ and $N_K=20$. (a) Test results for the 2D training (see text for details). (b) 
Test results for the 1D training (see text for details).}
\end{figure}

We proceed with testing how the CNN trained with standard map data performs in predicting chaos for other maps.
%We have shown that the neural network can be roughly divided into chaos regions using only short iterations. Then how does the network that trained only a standard map make predictions about other chaos systems?
We choose the logistic map as a simple one-dimensional chaotic test bed. The logistic map is written as $x_{n+1}=rx_{n}(1-x_{n})$. The parameter $r$ controls the crossover from regular to chaotic dynamics, which happens at $r_c \approx 3.56995$. 
We use two training methods. The first one trains the network only with the $p_n$ data sequence from
the standard map in Eq.~\ref{standard_map}. We coin that trained network 1D. The  second one is the original CNN discussed above, coined here 2D.  As shown in Fig.~\ref{logistic}, the network mainly generates errors at the boundary of chaos region similar to the standard map. For $2.5 \leq r \leq 4.0$ the accuracy is 84$\%$ for 2D network and 90\% for the 1D network.

\subsection{Training with the standard map, testing the Lorenz system}
\begin{figure} [ht]
	\includegraphics[width=\columnwidth]{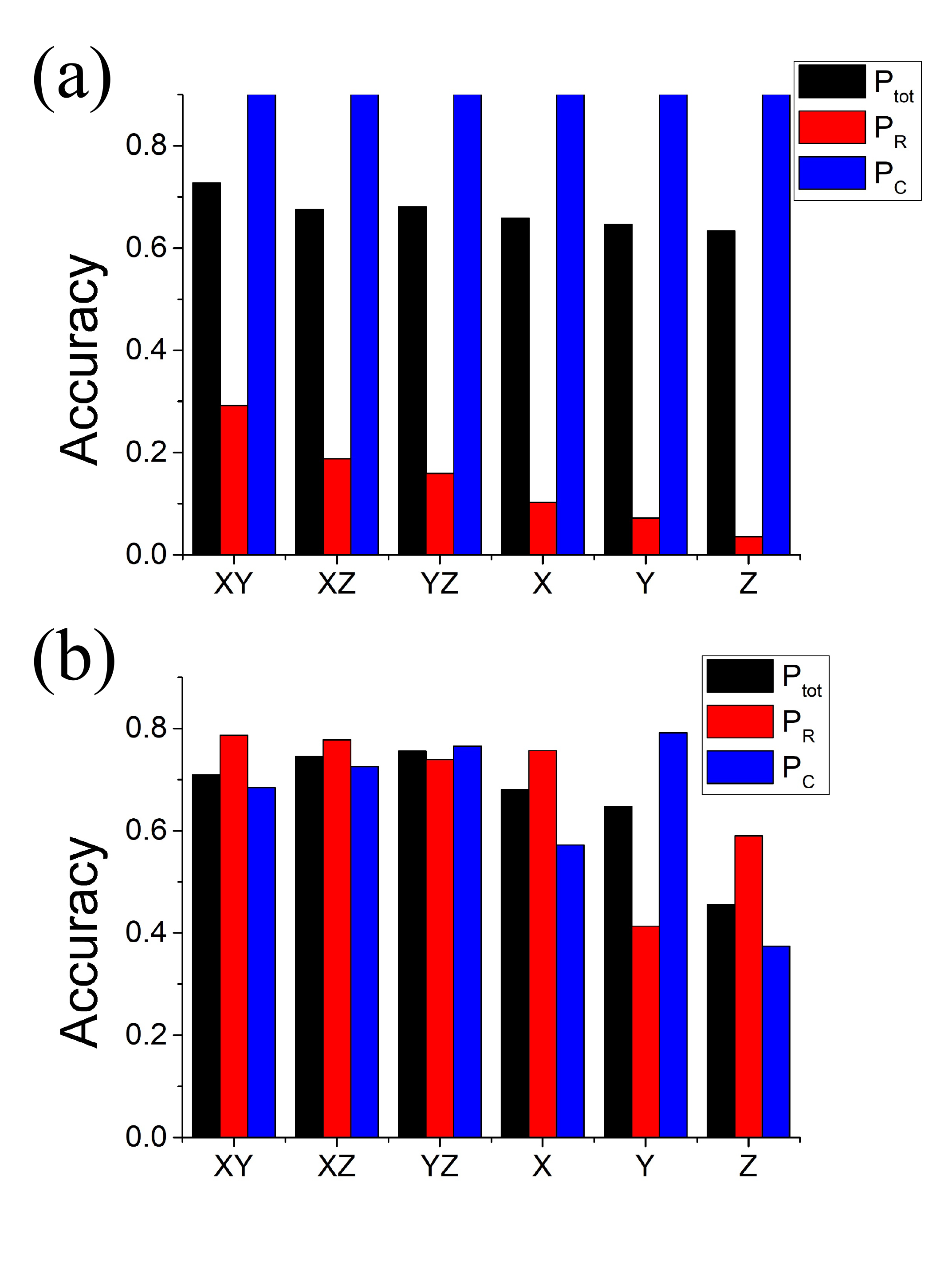}
	\caption{\label{lorenz} The result of predictions for the Lorenz system with a network trained from the standard map. 
%{\color{blue}{The $\rho$ range of the Lorenz system used in the test is 0 to 39.8.}} 
The XY, XZ, YZ bars represent the dimensions of the Lorenz system used as input to the network trained with $(p, x)$
data from the standard map. The training conditions are $N_K = 20$, $K_{min}=1.0$, $K_{max}=2.0$, and $M_{K}=11$. The X, Y, Z bars represent the single dimensions of the Lorenz system used as input to the network trained with $p$ data only from the standard map. (a) Accuracy without normalizing the trajectories of the Lorenz system. (b) Accuracy when normalizing trajectories of the Lorenz system. 
%{\bf \color{red} What are the parameters here? Training? Testing ($\rho$)?} {\bf \color{blue}{There is no training $\rho$. The network is trained with $N_K = 20$, $K_{min}=1.0$, $K_{max}=2.0$, and $M_{K}=11$ of the standard map, and the $\rho$ used in the test is $\rho = 0 \sim 39.8$ as written in the main text.}}
}
\end{figure}

Next we test Lorenz system which is a three-dimensional map, with a CNN trained on the two-dimensional standard map. The Lorenz system is given by the following map equations:
\begin{equation}
\begin{aligned} 
\label{lorenz_system}
X_{n+1}=X_{n}+\sigma \Delta(Y_n-X_n),\\
Y_{n+1}=Y_{n}+\rho \Delta X_{n}-\Delta X_{n}Z_{n}-\Delta Z_{n},\\ Z_{n+1}=z_{n}+\Delta X_{n}Y_{n}-\beta \Delta Z_n.
\end{aligned}
\end{equation}
The parameters $\sigma=10$, $\beta=\frac{8}{3}$, and $\Delta n =0.001$. The chaos parameter $0 \leq \rho \leq 39.8$ was varied in steps of
0.2.
Because the network is trained with 2D data (standard map), the prediction is performed by selecting only two dimensions in the 3D Lorenz system ($(X_n,Y_n), (X_n,Z_n), (Y_n,Z_n)$). As Fig.~\ref{lorenz} (a) shows, using trajectories obtained from Eq.~\ref{lorenz_system} directly as a network input classifies most of them as chaotic. We think this happens because the trajectory data of the standard map used for training are bounded between 0 and 1, but the trajectories from Lorenz system are not. Input values that exceed these boundaries cause nodes in the network to be active regardless of the input characteristics. Therefore we normalize the input data from the Lorenz system. This leads to a drastic
increase of accuracy as shown in Fig.~\ref{lorenz} (b). 
We also tested the outcome when selecting only one dimension in the Lorenz system for the input vector. We find
a strong reduction of the accuracy. We therefore conclude that the training and testing data are yielding best performance
when for both the minimum of the two dimensions (training map, testing map) is chosen.
%In the case of logistic maps, since the 1D network performance was better than the 2D network, we tested the 1D network in the lorenz system to see if this is a universal result. Contrary to the results of the logistic map, the accuracy decreased when only one dimension was selected. 
 
% {\color{red} Add the results on the logistic map, and on the Lorentz attractor. Place the figure environments at the text position of their first description. }

\section{Conclusion}
We trained convolutional neural networks with time series data from the two-dimensional standard map. As a result, the
network can classify unknown short trajectory sequences into chaotic or regular with high accuracy. 
To reach accuracies of up to 98\% we need trajectory segments with length less than 
5-10 Lyapunov times. Similar accuracies need 100-1000 longer segments when using traditional classifiers based
on measuring Lyapunov exponents. 
The main cause of errors is due to fractal phase space structures at the boundaries between chaotic and regular dynamics. Trajectories launched in these regions yield sticky trajectories which can mimick regular ones for long times,
only to escape at even larger times into the chaotic sea.
% is able to learn the chaotic features of the standard map and classify their time series with high accuracy. The vicinity of the boundary between regular and chaotic regions is often "sticky," that is, trapping orbits from the chaotic region for long times. These orbits cause errors in network evaluation. Sticky orbits look like regulars in a short time, so learning them as chaos can interfere with network training. From the point of view of the network, the orbits with the same characteristics may seem to have different labels. 
We also used a network trained with two-dimensional standard map data to classify chaotic and regular dynamics in
one- and three-dimensional maps. Surprisingly high accuracy is reached when the training data are projected into
one dimension for predictions on the one-dimensional logistic map, and when to-be-predicted data from the 
three-dimensional Lorenz system are projected onto two dimensions. We conclude that accuracy is optimized when
the minimum of the two dimensions (training map, testing map) is chosen for both training and testing.

\begin{acknowledgments}
This  work  was  supported  by  the  Institute  for  Basic  Science,  Project  Code  IBS-R024-D1.
SF thanks Konstantin Kladko for discussions during a visit to IBS, which led to the main idea of machine learning based chaos testing, and Natalia Khotkevych for early attempts to figure a realization pathway. 
\end{acknowledgments}

\appendix

\section{Details of the neural network structure}
\label{app1}

\begin{figure} [ht]
	\centering
	\includegraphics[width=\columnwidth]{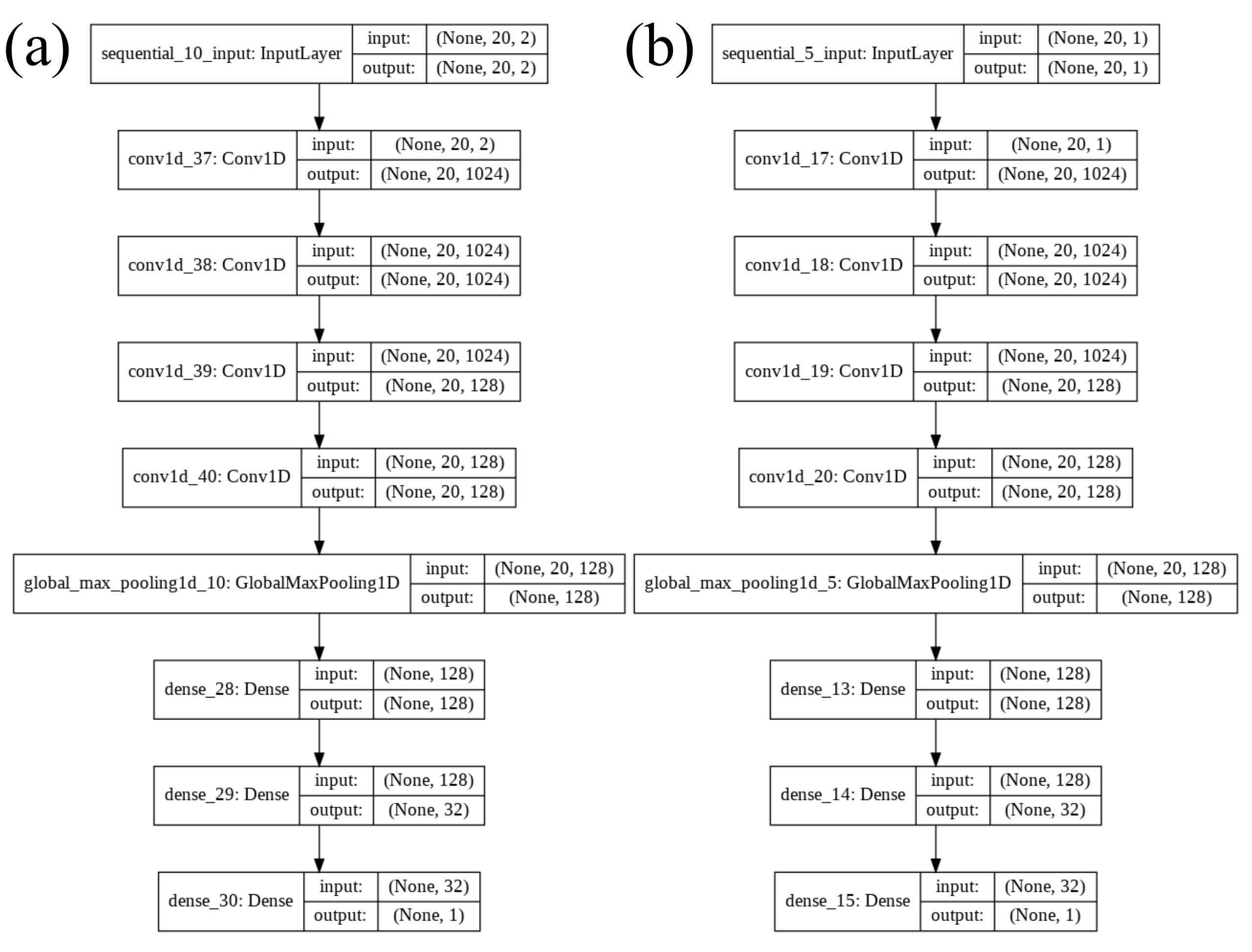}
	\caption{\label{fig_net_struct} Network architecture for chaos classification, consisting of four convolution layers, and three fully-connected layers. The filter size and output size are provided for each layer. Network structure for (a) 2D and (b) 1D shape training.}
\end{figure}

The neural network model we use to analyze the chaotic pattern is the convolutional neural network (CNN) \cite{LeCun} with a fully connected (FC) network \cite{krizhevsky,goodfellow2016deep}. The required nonlinear response of the system is provided by the rectified linear unit (ReLU) \cite{Nair,goodfellow2016deep}. For supervised learning, we use a cross entropy as loss function~\cite{goodfellow2016deep}. Fig. \ref{fig_net_struct} shows one of the CNN structures used here. In the figure, 1024 filters in the first layer scan the input data independently, then yield 1024 feature maps which are used as the input data for the second layer after applying the activation function ReLU.
After processing through all convolutional layers, we rearranged the pixels of the last feature maps into one-dimensional data for the fully connected layers. In the last layer we set the desired output: 1 is a chaotic, and 0 is a regular trajectory. \\

We use a two-dimensional convolutional filter. The data comprise multiple spatial channels and each channel gives time-series data. The relation between the 2D input vector $a$ and the convolution layer output vector $z$ is

\begin{equation} \label{eq.input_layer}
z^{(\ell)}_{i,j,m} = \sum^{F_{row}}_{q=1}\sum^{F_{col}}_{p=1}w^{(\ell)}_{p,q,m}a^{(\ell-1)}_{(i+p-1),(j+q-1)}+b^{(1)}_{m}.
\end{equation}

After calculating $z$, the nonlinear activation function (We used ReLU) is used to obtain the value of  $a$ of the next layer:
\begin{equation} \label{eq.activation}
a^{(\ell)}_{i,j,m} = \textrm{ReLU}(z^{(\ell)}_{i,j,m}),
\end{equation}
where $i, j$ are input element indices, $p, q$ are filter element indices, $m$ is the filter index (e.g. m = 1024 means that 1024 filters of size $F_{row} \times F_{col}$ were used), and $\ell$ is the layer index (the indices of input and first layer are $\ell$ = 0 and $\ell$ = 1 respectively). The weight $w^{(\ell)}_{p,q,m}$ is the $(p, q)^{th}$ element of $m^{th}$ filter. The bias $b_m$ is a constant. The filter size is $F_{row} \times F_{col}$, where we chose $F_{row}=2$ and $F_{col}=1$. To apply the filter to all the elements of the input, the boundary is filled with zeroes to match the size of the input, which is called zero padding~\cite{Dumoulin}. The 2D input through the convolution layer has a dimension of $(i, j, m)$ due to the number of filters $m$ in the convolution layer. Accordingly, the input / output relationship of the next layer is as follows:
\begin{equation} \label{eq.layer_layer}
z^{(\ell)}_{i,j,n} = \sum^{M_{\ell}}_{n=1}\sum^{F_{row}}_{q=1}\sum^{F_{col}}_{p=1}w^{(\ell)}_{p,q,n}a^{(\ell-1)}_{(i+p-1),(j+q-1),m}+b^{(\ell)}_{n},
\end{equation}
where $M_{\ell}$ is the number of filters between the $\ell^{th}$ and $(\ell-1)^{th}$ layers, $\ell$ has a range of 1 to $L$. After convolution, it processes through the activation function as shown in Eq.~\ref{eq.activation}. 

As shown in Fig.~\ref{fig_net_struct}, there is a pooling layer at the end of the convolution layers. This flattens the convolutional output by finding the maximum according to the filter dimension of the input (Eq.~\ref{eq.pooling}).

\begin{equation} \label{eq.pooling}
a^{(\ell)}_{m}=max(a^{(\ell)}_{i,j,m}).
\end{equation}
The reason for flattening the output is to use the convolutional output as the input of the fully connected layer:

\begin{equation} \label{eq.FCN_layer}
z^{(\ell)}_{i}=\sum^{K}_{j=1}w^{(\ell-1)}_{j,i}a^{(\ell-1)}_{j},
\end{equation}
where $w_{i,j}$ is a weighted connection between the $j^{th}$ component of $(\ell-1)^{th}$ layer and the $i^{th}$ component of $(\ell)^{th}$ layer and $K$ is the number of nodes in the $(\ell-1)^{th}$ layer. As in Eq.~\ref{eq.FCN_activation}, the activation function uses ReLU:
%, and the activation of the last layer uses softmax~\cite{goodfellow2016deep}.

\begin{equation} \label{eq.FCN_activation}
a^{(\ell)}_{i}=\textrm{ReLU}(z^{(\ell)}_{i}).
% \begin{cases}
% \textrm{ReLU}(z^{(\ell)}_{i}) & \ell < L \\
% \textrm{Softmax}(z^{(L)}_{i}) & \ell = L \\
% \end{cases}
\end{equation}

The output value of the network, when obtained in this way, is different from the desired output because it is obtained from the unfitted $w$ value. $w$ updates in the direction of reducing this difference and we call this difference the loss or cost. In this work, we selected the cross entropy as the loss function, defined as
\begin{equation}\label{eq.cost_fnc}
    C = -\sum_{k=1}^{N_{L}}{(a^{true}_{k}log(a^{(L)}_{k})+(1-a^{true}_{k})log(1-a^{(L)}_{k}))},
\end{equation}
where $N_{L}$ is the number of nodes in the output layer and $a^{true}_{k}$ is the desired output at the $k^{th}$ node. The cross-entropy loss function can reflect the degree of error to weight updates better than the mean square error loss function (MSE, MSE$=\frac{1}{N_{L}}\sum_{k=1}^{N_L}(a^{true}_k-a^{(L)}_k)^2$) because of the log term and is known as a cost function suitable for classification problems~\cite{nielsen2015neural}. Similar to the energy minimization problem in physics, supervised learning minimizes a cost function $C$ at the output layer.

Training the neural network means finding optimized parameters $w^{(\ell)}_{pqm}$ and $b_{n}^{(l)}$ that minimize $C$. We use an Adaptive Moment Estimation (Adam) algorithm for the optimization of learning~\cite{kingma2014adam}. As far as we know, the choice of optimization algorithm has little impact on the network performance, but is instead mainly related to the speed of learning. It is known that an Adam algorithm can find fitting variables faster than stochastic gradient descent methods, RMS prop and AdaDelta~\cite{kingma2014adam}.

% This is achieved by an iterative sequence where parameters are updated according to $\vec{v} = \vec{v} -\eta \nabla_{\vec{v}} C$ where ${\vec{v}}$ the parameter vector and $\eta$ is the learning rate parameter; we usually use $\eta = 0.001$. The gradient $\nabla_{\vec{v}}C$ is easily obtained by the back-propagation algorithm~\cite{Rumelhart2,Bouvrie}.

\section{Network comparison}
\label{app2}

In this section, we compare three different network architectures. 
First, we consider a fully connected neural network (FCN) and then compare it to other machine learning methods. 
The FCN is chosen because it is the most basic structure of deep learning classifiers. We also consider a recurrent neural network (RNN)~\cite{Sherstinsky2018}. The RNNs are usually used to deal with temporal dynamic behavior. 

We consider supervised classification with labels indicating chaos or regularity, where the input of the neural networks is fixed at $N_K=20$, and the output is one node for the corresponding label for training. 
The neural networks presented in this paper end with a sigmoid layer.

% \subsection{Fully connected network}
The first type of network considered in this section is a fully connected network, which consists of multiple fully connected layers and each layer has a nonlinear activation function. We use eight hidden layers with ReLU (Rectified Linear Unit) activation and the number of hidden neurons in each layer is [256, 256, 512, 512, 512, 256, 128, 64]. The network is trained with the Adaptive Moment Estimation (Adam) algorithm~\cite{kingma2014adam}.

% \subsection{Recurrent neural network}
Recurrent neural networks (RNN) are neural networks for processing sequential data. 
RNN uses the current input as well as any previously processed input.
This is possible with a loop structure between the RNN input and the output.
Each node in a given layer is connected with a directed connection to the current layer. 
Because of this, the RNN is expected to have a function of memory. 
The sequence itself has information, and recurrent networks use this information through the loop structure. 
We use three type of RNNs: simpleRNN~\cite{Sherstinsky2018}, LSTM~\cite{Sherstinsky2018,Gers1999}, and GRU~\cite{Cho2014}.
Three RNN cells(layers) were used, each with 200 hidden neurons. 
After the RNN cells, three fully connected layers are connected with the size of 200, 100 and 32 respectively.

\begin{table}[ht]
\centering
% \begin{tabular}{|c|c|c|c|c|c|c|c|}
\begin{tabular}{|c|ccc|} 
\hline
Classifiers & $P_{tot}$ & $P_{C}$ & $P_{R}$ \\\hline
% \hline
FCN        &  0.89  & 0.88  & 0.91 \\
SimpleRNN  &  0.94  & 0.96  & 0.92 \\
GRU        &  0.95  & 0.96  & 0.93 \\
LSTM       &  0.94  & 0.96  & 0.93 \\
CNN        &  0.96  & 0.94  & 0.97 \\
\hline
\end{tabular}

\caption{Performance for different deep learning classifiers. The networks are trained with chaotic and regular trajectories for $K_{min}=1.0,\;K_{max}=2.0,\;M_K=11,\;M_{tr}=2601$ and $N_K=20$. The network performances are evaluated for $K_{min}=3.0,\;K_{max}=3.5,\;M_K=6,\;M_{tt}=2601$ and $N_K=20$. For each K value, 2601 different initial values ($p_{0,i}, x_{0,j}$) were selected as ($p_{0,i}=(i-1)\frac{1}{50}, x_{0,j}=(j-1)\frac{1}{50},~ (i,j\in \mathbb{Z},\:1 \leq i,j \leq 51,\; )$). }
\label{network_comparison_table}
\end{table}

It is known that recurrent networks perform well for sequential data, but at least in our data sets there was no significant difference between using CNN and RNN.

\bibliography{ref}% Produces the bibliography via BibTeX.

\end{document}